\documentclass[10pt,twocolumn,letterpaper]{article}

\usepackage[pagenumbers]{cvpr} 

\usepackage{graphicx}
\usepackage{amsmath}
\usepackage{amssymb}
\usepackage{booktabs}

\usepackage{float}
\usepackage{algorithm}
\usepackage{algpseudocode}
\usepackage{algorithmicx}
\usepackage{amsfonts}
\usepackage{bbm}

\usepackage{multirow}
\usepackage{makecell}
\usepackage{threeparttable}

\usepackage[pagebackref,breaklinks,colorlinks]{hyperref}

\usepackage[capitalize]{cleveref}
\crefname{section}{Sec.}{Secs.}
\Crefname{section}{Section}{Sections}
\Crefname{table}{Table}{Tables}
\crefname{table}{Tab.}{Tabs.}

\begin{document}

\title{SNF: Filter Pruning via Searching the Proper Number of Filters}
\author{Pengkun Liu$^1$, Yaru Yue$^1$\thanks{Equal contribution.}, Yanjun Guo$^1$, Xingxiang Tao$^1$, Xiaoguang Zhou$^{1}$\thanks{Corresponding author.} \\
$^1$School of Modern Post (School of Automation), Beijing University \\ of Posts and Telecommunications, Beijing, China.\\
{\tt\small \{pkl, yyr, gyj811, taoxx, zxg\}@bupt.edu.cn}
}
\maketitle

\begin{abstract}

Convolutional Neural Network (CNN) has an amount of parameter redundancy, filter pruning aims to remove the redundant filters and provides the possibility for the application of CNN on terminal devices. However, previous works pay more attention to designing evaluation criteria of filter importance and then prune less important filters with a fixed pruning rate or a fixed number to reduce convolutional neural networks' redundancy. It does not consider how many filters to reserve for each layer is the most reasonable choice.
From this perspective, we propose a new filter pruning method by searching the proper number of filters (SNF). SNF is dedicated to searching for the most reasonable number of reserved filters for each layer and then pruning filters with specific criteria. It can tailor the most suitable network structure at different FLOPs. 
Filter pruning with our method leads to the state-of-the-art  (SOTA) accuracy on CIFAR-10 and achieves competitive performance on ImageNet ILSVRC-2012.
SNF based on the ResNet-56 network achieves an increase of 0.14\% in Top-1 accuracy at 52.94\% FLOPs reduction on CIFAR-10. Pruning ResNet-110 on CIFAR-10 also improves the Top-1 accuracy of 0.03\% when reducing 68.68\% FLOPs. For ImageNet, we set the pruning rates as 52.10\% FLOPs, and the Top-1 accuracy only has a drop of 0.74\%. The codes can be available at https://github.com/pk-l/SNF.

\end{abstract}


\section{Introduction}
\label{sec:intro}

Convolutional Neural Network (CNN) has attained great success in computer vision, like image classification \cite{russakovsky2015imagenet, chollet2017xception}, object Detection \cite{lin2014microsoft, ren2015faster} and semantic segmentation \cite{long2015fully, cordts2016cityscapes}. 
However, the computing resource consumption rises rapidly as the models become more complex. Lots of float point operations (FLOPs) and memory footprint have caused resistance to the general application in edge devices, like wearable devices and embedded devices, etc.
To compress and accelerate convolutional neural networks, the approaches emerge endlessly. These approaches can be classified into five categories:  structured pruning \cite{li2016pruning,li2020group,he2019filter,he2017channel}, unstructured pruning \cite{han2015deep, ding2019global,guo2016dynamic,han2015learning}, parameter quantization \cite{ba2013deep,banner2018post,liu2018bi}, knowledge distillation \cite{hinton2015distilling,liu2019structured,yim2017gift} and low-rank decomposition \cite{denton2014exploiting,jaderberg2014speeding}.

\begin{figure}[t]
  \centering
   \includegraphics[trim=0 0 0 0,scale=0.6]{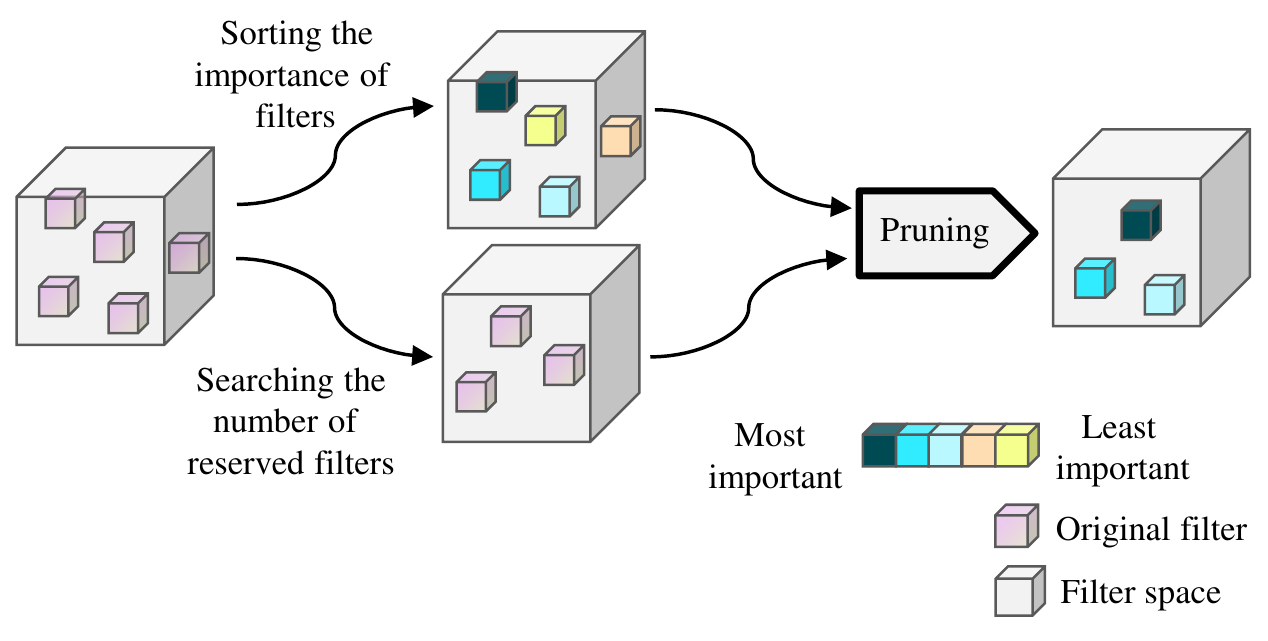}
   \caption{The pipeline of SNF. Our method has two branches: sorting the importance of filters and searching the number of reserved filters. 
}
   \label{fig:pip}
  \vspace{-0.4cm}
\end{figure}

Channel pruning is one of the structured pruning methods, which attracted the attention of researchers for its implementation-friendly. In terms of the different operation bases, channel pruning \cite{he2017channel,he2019filter} can be divided by filter pruning and feature map pruning. No matter which method ends up removing the channels. 
In this paper, we only focus on filter pruning. Filter pruning slims the network to reduce the float point operations (FLOPs) and memory footprint by removing the number of filters in convolutional layers. There are two advantages that it is friendly to operate for any convolutional neural network structure and it can produce a thinner model without special devices. 
However, the ability of the model representation is limited to the network width. It is challenging to prune networks without the accuracy drop. 

Existing filter pruning methods usually remove less important filters relatively by designing criteria to evaluate filter importance, like Channel Pr \cite{li2016pruning}, FPGM \cite{he2019filter}, Slim \cite{liu2017learning} and APoZ \cite{hu2016network}, etc.
These methods do not consider how many filters should be removed in each layer and set the pruning rates manually. 

Inspired by \cite{ye2018rethinking, liu2019metapruning}. we claim that the number of filters pruned in each layer is also critical to the model performance. For each layer, it has the most suitable number of filters at a certain FLOPs reduction.
For a convolutional layer, the representation capability is determined by the number of filters in this layer and the ability of filter representation.
According to \cite{howard2017mobilenets}, the ability of the network representation is limited to the network width. So we believe that the accuracy drop after pruning is related closely to the number of pruned filters.
On the other hand, we theoretically analyze that the ability of filter representation is close to each other in the same layer. 

Based on our analysis above, we propose a method by searching for the proper number of filters (SNF). Our SNF can search for the most suitable number of reserved filters for each layer automatically, and then prunes filters with a criterion of filter importance ranking. 

Our main contributions are summarized as follows.
\begin{itemize}
\item We theoretically analyze that the ability of filter representation is close to each other in the same layer. Unlike other papers mentioned, we consider that not only the criteria of filter importance ranking is important, but the number of reserved filters is also equally important or even more significant. 

\item We propose a new method called SNF, which consists of searching the most reasonable number of reserved filters for each layer and pruning filters with specific criteria. Our approach can tailor the most suitable number of filters for each layer at different FLOPs reduction.

\item We verify our approach on CIFAR-10 and ImageNet using ResNet-56/110 and ResNet-50 respectively. The experiment results have proved that our method achieves the state-of-the-art performance on CIFAR-10 and has a quite competitive accuracy on ImageNet.

\end{itemize}

\section{Related work}
\label{sec:related}
In this section, we mainly discuss the existing pruning methods, including structured pruning, unstructured pruning methods.

\subsection{Structured Pruning}
Structured pruning tries to remove the channels or layers. As we know, many criteria \cite{li2016pruning,he2019filter,lin2020hrank,liu2017learning,molchanov2016pruning} are designed to evaluate which channel is less important and remove them. 
Channel Pr \cite{li2016pruning} took $L_1$-norm as a criterion for sorting the importance of filters, which prune filters with the smaller norm.
FPGM \cite{he2019filter} removed the most replaceable filters containing redundant information based on the criteria of the geometric median.
Thinet \cite{luo2017thinet} treated channel pruning as an optimization problem. This method uses a greedy strategy to extract the optimal feature subset of the latter layer's input features, and then the channels corresponding to the feature subset were reserved.
HRank \cite{lin2020hrank} computed the rank of feature maps. This method considered the feature maps with low rank contained less information. Pruning the channels corresponding to the low-rank feature maps had less impact on the model performance.
As the Batch Normalization (BN) layer is widely used, Slim \cite{liu2017learning} proposed to apply L1 regularization on BN layers to make the scaling factor sparse, and then the channels with small scaling factor were removed.
Soft filter pruning \cite{he2018soft} reset the less important filters but not removed these filters.
With reinforcement learning, AMC \cite{he2018amc} took accuracy as a reward function to prune the network.
Metapruning \cite{liu2019metapruning} utilized meta-learning to train a pruning network which provides weights for all the possible sub-networks, and then searched for the best pruned network structures.
Pavlo et al. \cite{molchanov2019importance} used the Taylor expansions to approximate a filter's contribution and then took the first item as the criteria of pruning.
C-SGD \cite{ding2019centripetal} trained several filters to collapse into a single point in the parameter hyperspace and discard all but leave one filter.
Gao et al. \cite{gao2021network} proposed the problem of loss-metric mismatch, and this method designed a performance prediction network to make performance maximization. 
ResRep \cite{ding2021resrep}  proposed to re-parameterize CNN into the maintain parts and prune parts. 
Zi et al. \cite{wang2021convolutional} analyzed the network pruning from a perspective of redundancy and then removed redundant filters by building a redundancy graph. 

\subsection{Unstructured Pruning}
The previous works \cite{denil2013predicting, frankle2018lottery} have proved that convolutional neural network (CNN) has the parameter redundancy and the accuracy of the model does not drop while most of the neurons are removed. Unstructured pruning proposed that removing most of the weights does not affect model performance. 
Han et al. \cite{han2015learning} pruned less important weights with the criteria of $L_1$-norm and $L_2$-norm.
SNIP \cite{lee2018snip} utilized the gradient of the loss function with respect to the weight connection to reset small gradient weights. 
These methods only make the matrix sparse, but the model will not be compressed and accelerated if we have no special devices. 
\begin{figure*}
  \centering
  \begin{subfigure}{1.\linewidth}
  \centering
    \includegraphics[scale=0.7]{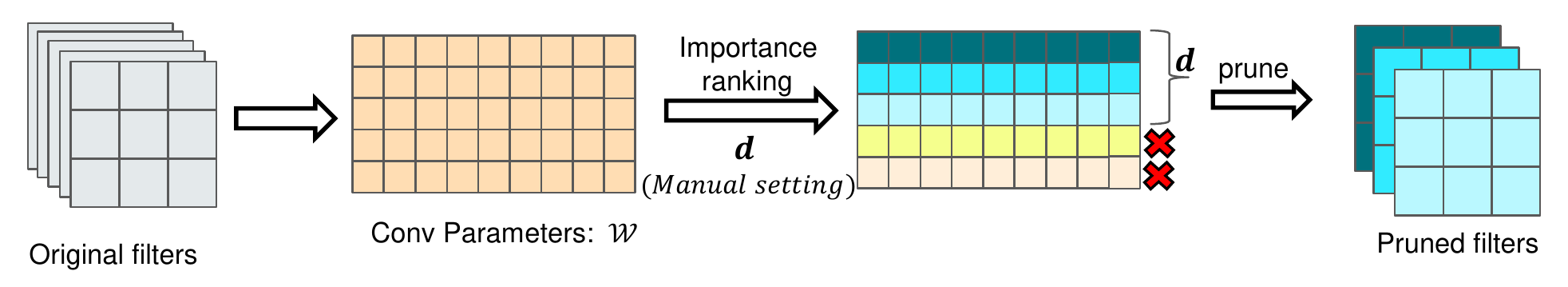}
    \caption{Traditional process.}
    \label{fig:process-a}
  \end{subfigure}
  -------------------------------------------------------------------------------------------------------\\
  \begin{subfigure}{1.\linewidth}
  \centering
    \includegraphics[trim=0 0 0 0,scale=0.7]{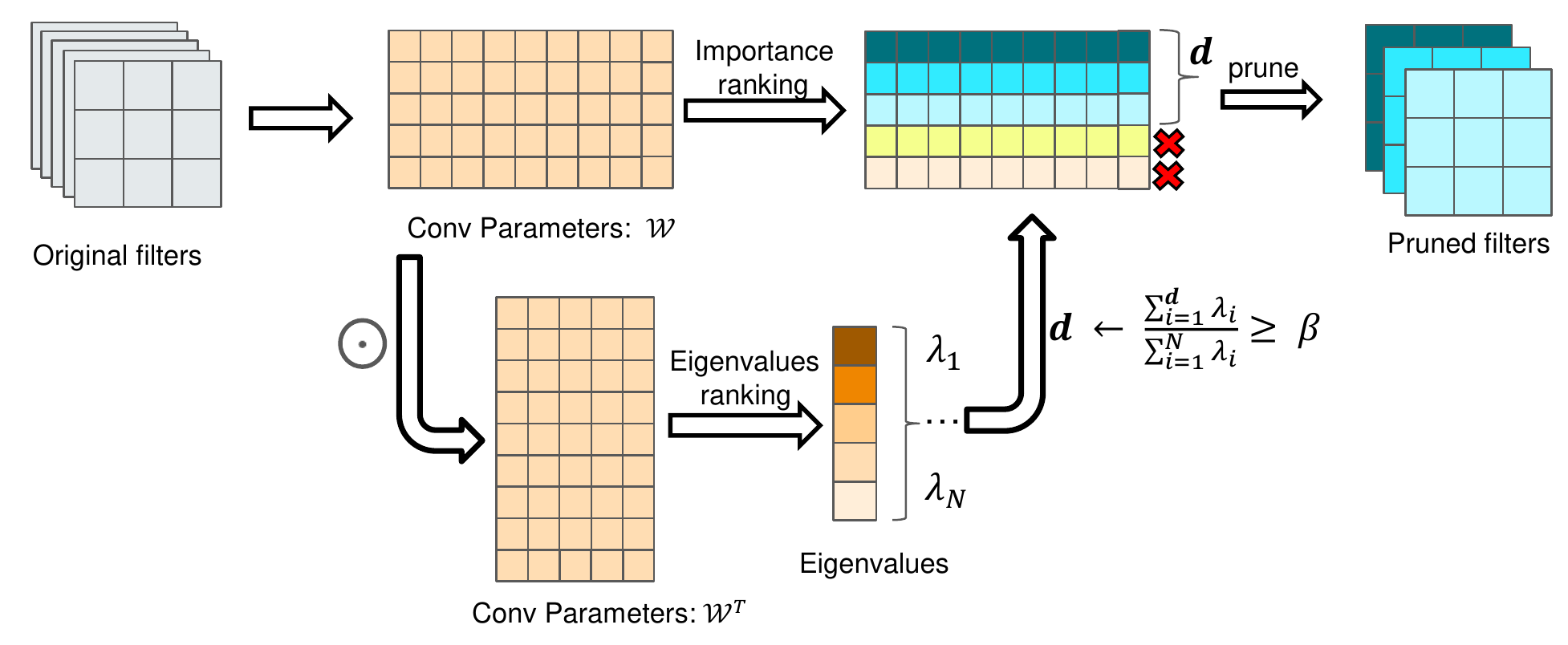}
    \caption{SNF.}
    \label{fig:process-b}
  \end{subfigure}
      \caption{Traditional process $vs$ SNF. We visualize the process of pruning a $3 \times 3$ convolutional layer with 5 filters. For a kernel $ \mathcal{K} \in \mathcal{R}^{3 \times 3} $, we flatten it into $ \mathcal{W} \in \mathcal{R}^{1 \times 9}$. (a) Traditional process. Traditional methods always prune filters with a special criterion and set the number of reserved filters manually. (b) SNF. Our method also uses a specific criterion to sort the filter importance, but the number of filters reserved is adjusted automatically according to Algorithm \ref{alg:SNF}. 
  }
  \label{fig:short}
\end{figure*}

\section{Methodology}
\label{sec:method}
\subsection{Notations}
To describe our method clearly, let us first introduce necessary notations. For the convolutional neural network, we define the number of layers as $L$. For $l$-th layer with $N$ filters ($N$ is the number of the filters), we take the $ \{\mathcal{X}_1, \mathcal{X}_2,..., \mathcal{X}_N \} $ as the ability of filter representation in this layer. Let $ \{\mathcal{D}(\mathcal{X}_i), i = 1,2,...,N\}$ be the variance contribution of filters for this layer.
For $l$-th layer, the $N_{l-1}$ and $\mathcal{F}_{l-1}$ represent the number of input channels and input feature maps respectively. The number of output channels and output feature maps are represented by $N_{l}$, $F_{l}$, and the output feature map size is $h_{l} \times w_{l}$. The convolutional layer parameters can be defined as $\{\mathcal{W}^{(l)} \in \mathcal{R}^{n_{l} \times n_{i-l} \times h_{l} \times w_{l}},\ l = 1,2...,L\}$. 

\subsection{Ability of Filter Representation} 
In this section, we attempt to introduce the ability of filter representation in the same layer that is close to each other. For $l$-th layer, we naturally assume that the ability of the filter representation is positive, but the filter contribution for the layer representation cannot be infinite. In other words, the variance contribution of filters has an upper limit. We have
\begin{equation}\label{eq-con-upper}
    \begin{split}
        & \quad\quad\quad\quad \exists \ \epsilon > 0,\ s.t.\ \mathcal{X}_{i} > \epsilon,\\
        & \exists \ k > 0,\ s.t.\ D(\mathcal{X}_{i}) <= k. \quad \forall i \in \{1,2,...,N\}.
    \end{split}
\end{equation}

According to Chebyshev’s inequality, we can get
\begin{equation}\label{eq-chebyshev}
    P(\frac{1}{N} \sum_{i=1}^N | \mathcal{X}_i - E(\mathcal{X}_i)| \geq b) \leq \frac{Var(\sum_{i=1}^N \mathcal{X}_i)}{b^2},
\end{equation}
where $E(\mathcal{X}_i)$ is the average of the ability of filter representation, $b$ is a positive number, $Var(\sum_{i=1}^N \mathcal{X}_i)$ is the filter variance contribution in the layer. In terms of the variance formula, we have
\begin{equation}\label{eq-var-dec}
\begin{split}
    Var(\sum_{i=1}^N \mathcal{X}_i) &= \frac{1}{N^{2}} Var(\sum_{i=1}^{N} \mathcal{X}_{i}) \\
    &= \frac{1}{N^{2}} (\sum_{i=1}^{N} D(\mathcal{X}_i) + \sum_{i \ne j} Cov(\mathcal{X}_i, \mathcal{X}_j)), \\
\end{split}
\end{equation}
where $i,j \in \{1, 2, ..., N\}$. For $l$-th layer, some filters are weakly correlated to each other. 
We further define that there are $\zeta N (0< \zeta <1)$ weakly correlated filters in the $l$-th layer, we have
\begin{equation}\label{eq-cov-inej}
    \sum_{i \ne j} Cov(\mathcal{X}_i, \mathcal{X}_j) <= \zeta N k,\quad i,j \in \{1, 2, ..., N\}.
\end{equation}

By Equation \ref{eq-chebyshev}, \ref{eq-var-dec} and \ref{eq-cov-inej}, we can get

\begin{equation}\label{eq-final}
    P(\frac{1}{N} \sum_{i=1}^N | \mathcal{X}_i - E(\mathcal{X}_i)| \geq b) \leq \frac{k(1+\zeta)}{b^2 N}.
\end{equation}

According to the Equation \ref{eq-final}, the $P(\frac{1}{N} \sum_{i=1}^N | \mathcal{X}_i - E(\mathcal{X}_i)| \geq b)$ is close to 0 in possibility while $N$ is large enough. In other words, almost all the $\mathcal{X}_i$ are near the average. The ability of filter representation is close to each other. So we can further ensure the ability of layer representation is limited to the number of reserved filters in the layer. The above conclusion has an assumption that $N$ is large enough. In fact, the assumption can be relaxed in the real world, we provide evidence to validate the analysis in our ablation study. Above all, it is the key for the pruned model performance to search for the proper number of reserved filters in each layer.

\subsection{Searching for the Proper Number of Filters}

The ability of filter representation in the same layer is close to each other. We consider searching for the proper number of filters to reconstruct the original filter parameters. 
Searching the number of reserved filters can be defined as an optimization problem. For $l$-th layer, each filter is represented by a feature vector $W_i, 0<i<N$, where $N$ is the number of filters in this layer. 
We try to use the $d$ filters to reconstruct the $N$ filters. At the same time, the error between them is as little as possible. In fact, the smaller the reconstruction error is, the easier it is to reconstruct the original filter parameters. We refer to the Principal component analysis thought which reduces the high-dimensional data to low-dimensional to search for the proper number of filters reserved in this layer. The reconstruction error can be defined as the mean square error between reduced dimensionality $\hat{\mathcal{W}}$ and original $ \mathcal{W}$. The formula can be represented as

\begin{equation}\label{eq-reconstruction}
\begin{split}
    \sum_{i=1}^N||\mathcal{W}_i - \hat{\mathcal{W}_i}||_2^2 &= \sum_{i=1}^N||\mathcal{W}_i - \sum_{j=1}^d u_j^T \mathcal{W}_i u_j ||_2^2 \\ 
    &= -tr(\mathbf{U}^T (\sum_{i=1}^n \mathcal{W}_i \mathcal{W}_i^T) \mathbf{U})
\end{split}
\end{equation}
where $ \mathbf{U} $ is the projection matrix, $ i \in \{1,2,...,N\}$ and $ j \in \{ 1,2...,d \}$. Then the optimization formula can be defined as
\begin{equation}\label{eq-reconstruction-err}
    \underset{\mathbf{U}}{min} -tr(\mathbf{U}^T (\sum_{i=1}^n \mathcal{W}_i \mathcal{W}_i^T) \mathbf{U}) \quad s.t.\ \mathbf{U}^T \mathbf{U} = \mathbf{I}.
\end{equation}
To solve Equation \ref{eq-reconstruction-err}, we utilize Lagrange Multiplier Method,
then we can get
\begin{equation}\label{eq-lagrange-sim}
    \mathcal{W} \mathcal{W}^T u_i = \lambda_i u_i
\end{equation}
Finally, we set a reconstruction threshold of $\beta$, and the $\beta$ is equal to $1- \sum_{i=1}^N||\mathcal{W}_i - \hat{\mathcal{W}_i}||_2^2$. Let d be the number of reserved filters. So we can use the following formula to search how many filters to keep.
\begin{equation}\label{eq-reconstruction-thr}
    d \leftarrow 
    \frac{\sum_{i=1}^d \lambda_i}{\sum_{i=1}^N \lambda_i}  \geq \beta.
\end{equation}

\subsection{Reconstruction Threshold}
The number of reserved filters depends on the reconstruction threshold $\beta$, and the $\beta$ is determined by the FLOPs. The reconstruction threshold $\beta$ can be adjusted by Binary search to fit the FLOPs we need.
For $l$-th layer, suppose the reconstruction threshold $\beta$, then $(N_l - d_l)$ filters should be removed according to Equation \ref{eq-reconstruction-thr}. After pruning, we can compute the pruning ratio by comparing the current FLOPs (after pruning) with the original FLOPs (before pruning). If the current pruning ratio can not meet our requirement, the reconstruction threshold $\beta$ will adjust automatically. The process description of SNF can refer to Algorithm \ref{alg:SNF}.

\renewcommand{\algorithmicrequire}{\textbf{Input:}}  
\renewcommand{\algorithmicensure}{\textbf{Output:}} 

\begin{algorithm}[t]
    \caption{Algorithm Description of SNF.}
	\label{alg:SNF}
	\begin{algorithmic}[1]
        \Require
            $\mathcal{W}$ : Convolutional parameters, 
            $\theta $ : Pruning ratio,
            $\beta$ : reconstruction threshold,
            $F_c$ : the current FLOPs,
            $F_o$ : the original FLOPs;
        \Ensure
            $d$ : the number of reserved filters;
        \State set $\beta = rand(0,1)$
        \While{ $ 1 - F_c/F_o \leq \theta$ }
    	    \For{$l=1$; $l\leq L$; $l++$}
	            \State center the convolutional parameters: $\mathcal{W} \leftarrow \frac{1}{N} \sum_{i=1}^{N} \mathcal{W}_i $;
	            \State compute and sort the eigenvalue $\lambda$ of Covariance Matrix $ \mathcal{W}\mathcal{W}^T $;
	            \For{$j=1; j \leq n; j++$}
	                \If{$\frac{\sum_{i=1}^j \lambda_i}{\sum_{i=1}^N \lambda_i} >= \beta$}
	                \State $d_l = j$;
	                \EndIf
	            \EndFor
	        \EndFor
	        \State update the $\beta$ according to the $F_c/F_o$ and $\theta$;
	       \EndWhile
	       \State prune filters with the $L_1$-norm criteria
	\end{algorithmic} 
\end{algorithm}

\section{Experiments}
\label{sec:exp}

In this section, we evaluate SNF on CIFAR-10 \cite{krizhevsky2009learning} and ImageNet ILSVRC-2012\cite{deng2009imagenet} using ResNet. We use the one-shot pruning scheme.
For a fair comparison, all experiments are based on pycls \cite{radosavovic2019network} and retain data augmentation including random cropping and flipping.

\subsection{Datasets and Evaluation}
\textbf{Datasets}.
Our experiments are based on the datasets of CIFAR-10 \cite{krizhevsky2009learning} and ImageNet ILSVRC-2012 \cite{deng2009imagenet}. We use ResNet-56/110 on CIFAR-10 and ResNet-50 \cite{he2016deep} on ImageNet. For CIFAR-10, it consists of 10 classes images with $32 \times 32$ resolution, which contains 50,000 and 10,000 colorful images for training and testing respectively. For ImageNet, it is composed of 1.28 million training images and 50,000 testing images, which includes 1,000 classes natural images with $224 \times 224$ resolution.

\textbf{Evaluation Protocols}.  
For a fair comparison, we use floating-point operations per second (FLOPs) to evaluate model complexity. As the same as other literature, we use Top-1 and Top-5 accuracy on ImageNet, and Top-1 accuracy on CIFAR-10.

\subsection{Implementation Details}
\textbf{CIFAR-10.}
\label{res56-setting}
For CIFAR-10 dataset, the baseline models of ResNet-56/110 are trained from scratch (initial training) with a batch size of 64, the initial learning rate of 0.1, and a weight decay of $5e-4$. Furthermore, we use stochastic gradient descent (SGD) with a momentum of 0.9 and train for 200 epochs. With the step schedule, the learning rate is divided by 10 at the epochs 60, 120, 160. After pruning, we fine-tune 200 epochs for the pruned model with an initial learning rate of 0.01. Other settings have no changes compared with initial training.

\textbf{ImageNet.}
For ImageNet ILSVRC-2012, we train ResNet-50 from scratch for 100 epochs, using SGD with a momentum of 0.9. The weight decay is set to $ 1e-4 $. The initial learning rate is equal to 0.2 with a batch size of 256. The policy of learning rate decay adopts the half-period cosine annealing schedule. After pruning, the settings of fine-tuning are the same as training in addition to the initial learning rate divided by 10.  

\textbf{Reconstruction Threshold.}
In our experiments, we take the $ \beta $ as the global reconstruction threshold. In fact, the reconstruction threshold for each layer must be larger than or equal to the global threshold $ \beta$. The reconstruction threshold $ \beta $ does not require to be set manually. The pruning ratio we need once is determined, the global reconstruction threshold $ \beta $ will adjust by pruning ratio automatically. The least number of filters for reconstructing the original filters can be searched referring to $ \beta $. No more extra hyper-parameters except the pruning ratio needs to be set manually.

\subsection{Pruning Results on CIFAR-10 and ImageNet}
We compare our approach to the representative methods. For a fair comparison, we use ResNet-56/110 and ResNet-50 on CIFAR-10 and ImageNet respectively since most of the previous state-of-the-art methods are based on them.

\textbf{Pruning ResNet-56 on CIFAR-10.} For the ResNet-56 network, we compared our method with \cite{yu2018nisp, he2017channel,he2018amc, he2019filter, he2018soft,he2020learning, ding2021resrep, lin2020hrank, xu2020trp}. Table \ref{table-res56-cifar10} shows the pruning results of ResNet-56 on CIFAR-10. We try to evaluate our approach at different pruning rates. The results show our method achieves the state-of-the-art performance on ResNet-56 in comparison to previous works. As shown in Table \ref{table-res56-cifar10}, we train ResNet-56 from scratch with the setting as section \ref{res56-setting}, the accuracy of the baseline model is 93.61\%. After pruning and fine tuning, SNF achieves 94.13\% Top-1 accuracy which increases by 0.14\% when reducing 52.94\% FLOPs. In terms of Top-1 accuracy drop, our method outperforms ResRep\cite{ding2021resrep} by 0.14\% and  LFPC\cite{he2020learning} by 0.49\% at the similar FLOPs reduction. Compared with NISP \cite{yu2018nisp}, Channel Pr \cite{he2017channel}, AMC \cite{he2018amc}, FPGM \cite{he2019filter}, SFP \cite{he2018soft}, we not only prune the model with a larger FLOPs reduction, but also the accuracy of pruned model is higher than them. For the FLOPs reduction of 77.93\%, the accuracy of our method is only 0.82\% drop. In comparison with ResRep \cite{ding2021resrep}, HRank \cite{lin2020hrank}, TRP \cite{xu2020trp}, our method still attains the best performance.
To our best knowledge, the Top-1 accuracy drop of our pruned model outperforms the results reported elsewhere at 52.94\% and 77.93\% FLOPs reduction. In fact, not only the Top-1 accuracy drop is better than the performance with other pruning approaches, but the Top-1 accuracy of the pruned model also outperforms others. 
\begin{table}[t]
    \caption{Pruning results of ResNet-56 on CIFAR-10. "Base Top-1" and "Pruned Top-1" indicate the Top-1 accuracy of the baseline model and pruned model. "Top-1 $\downarrow$" denotes the Top-1 accuracy drop between pruned model and the baseline model. The "FLOPs $\downarrow$" denotes the reduction of computations. Results marked with ’-’ are not reported by the authors.}
    \label{table-res56-cifar10}
 	\vspace{-0.5cm}
	\setlength{\tabcolsep}{0.2em}
	\begin{center}
		\begin{tabular}{l|cccc}	
            \toprule[1.5pt]
            
			\makecell[l]{Method}  &\makecell[c]{Base \\ Top-1(\%)}    &\makecell[c]{Pruned \\ Top-1 (\%)}	&\makecell[c]{Top-1 \\ $\downarrow$ (\%)} 	&\makecell[c]{FLOPs \\ $\downarrow$ (\%)}\\
			\midrule[1pt]
			
			NISP \cite{yu2018nisp}      &-      &-          &0.03      &42.6   \\
			Channel Pr\cite{he2017channel}    &92.8   &91.8   &1.0    &50 \\
			AMC \cite{he2018amc}     &92.8  &91.9      &0.9       &50   \\
			FPGM \cite{he2019filter}     &93.59  &93.26      &0.33       &52.6   \\
			SFP \cite{he2018soft}     &93.59  &93.35      &0.24       &52.6   \\
			LFPC \cite{he2020learning}     &93.59  &93.24      &0.35       &52.9   \\
			ResRep \cite{ding2021resrep}     &93.71  &93.71      &0.00       &52.91   \\
			\textbf{SNF(ours)}   &\textbf{93.61}  &\textbf{93.75}  &\textbf{-0.14}   &\textbf{52.94}   \\
			HRank \cite{lin2020hrank}   &93.26  &90.72      &2.54       &74.1   \\
			TRP \cite{xu2020trp}     &93.14  &91.62      &1.52       &77.82   \\
			ResRep \cite{ding2021resrep}     &93.71  &92.66      &1.05       &77.83   \\
			\textbf{SNF(ours)}   &\textbf{93.61}  &\textbf{92.79}  &\textbf{0.82}   &\textbf{77.93}   \\
			
			\bottomrule[1.5pt]
		\end{tabular}
	\end{center}
\end{table}
\begin{table}[ht]
    \caption{Pruning results of ResNet-110 on CIFAR-10. All the labels have the same meaning with Table \ref{table-res56-cifar10}.}
    \label{table-res110-cifar10}
 	\vspace{-0.5cm}
	\setlength{\tabcolsep}{0.2em}
	\begin{center}
		\begin{tabular}{l|cccc}	
            \toprule[1.5pt]
            
			Method  &\makecell[c]{Base \\ Top-1(\%)}    &\makecell[c]{Pruned \\ Top-1 (\%)}	&\makecell[c]{Top-1 \\ $\downarrow$ (\%)} 	&\makecell[c]{FLOPs \\ $\downarrow$ (\%)}\\
			\midrule[1pt]
			
			Li et al. \cite{li2016pruning}    &93.53   &93.30   &0.23    &38.60 \\
			SFP \cite{he2018soft}     &93.68  &93.86      &-0.18      &40.8   \\
			NISP-110 \cite{yu2018nisp}      &-      &-          &0.18      &43.78   \\
			GAL-0.5 \cite{lin2019towards}     &93.50  &92.74      &0.76       &48.5   \\
			FPGM \cite{he2019filter}     &93.68  &93.74      &-0.06       &52.3   \\
			ResRep \cite{ding2021resrep}     &94.64  &94.62      &0.02       &58.21   \\
			\textbf{SNF(ours)}   &\textbf{93.93}  &\textbf{94.23}  &\textbf{-0.30}   &\textbf{58.32}   \\
			LFPC \cite{he2020learning}     &93.68  &93.07      &0.61       &60.3  \\
			C-SGD \cite{ding2019centripetal}   &94.38  &94.41      &-0.03   &60.89   \\
			HRank \cite{lin2020hrank}   &93.50  &92.65      &0.85       &68.6   \\
			\textbf{SNF(ours)}   &\textbf{93.93}  &\textbf{93.96}  &\textbf{-0.03}   &\textbf{68.68}    \\
			
			\bottomrule[1.5pt]
		\end{tabular}
	\end{center}
\end{table}

\textbf{Pruning ResNet-110 on CIFAR-10.}
In Table \ref{table-res110-cifar10}, we report the pruning results of ResNet-110 on CIFAR-10. With the deeper network, we compare our method with \cite{li2016pruning, he2018soft, yu2018nisp, lin2019towards, he2019filter, ding2021resrep, he2020learning, ding2019centripetal, lin2020hrank}. As shown in Table  \ref{table-res110-cifar10}, our approach also achieves the state-of-the-art  performance. We train ResNet-110 from scratch and achieve the Top-1 accuracy of 93.93\%. Compared to initial training, the Top-1 accuracy of pruned model increases by 0.30\% at 58.32\% FLOPs reduction. Under the similar pruning ratio, the Top-1 accuracy with our method is 0.32 \% higher than RepRep \cite{ding2021resrep}. In comparison with Li ea al. \cite{li2016pruning}, SFP \cite{he2018soft}, NISP-110 \cite{yu2018nisp}, GAL-0.5 \cite{lin2019towards}, FPGM \cite{he2019filter}, SNF still outperforms them at higher FLOPs reduction. For GAL-0.5 \cite{lin2019towards} and FPGM \cite{he2019filter}, we achieve less Top-1 accuracy drop (-0.30\% $vs$ 0.76\% and -0.06\%) while the FLOPs reduction is larger (58.32\% $vs$ 48.5\% and 52.3\%). When pruning 68.68\% FLOPs with our method, the Top-1 accuracy increases by 0.03\%. Compared with HRank \cite{lin2020hrank}, our method outperforms by 0.88\% at a higher pruning ratio(68.68\% $vs$ 68.6\%). At the same accuracy drop, the FLOPs reduction with our approach is 7.79\% higher than C-SGD \cite{ding2019centripetal} (68.68\% $vs$ 60.89\%).

\begin{table*}
	
    \caption{Pruning results of ResNet-50 on ILSVRC-2012. "Base Top-5" and "Pruned Top-5" indicate the Top-5 accuracy of the baseline model and pruned model. "Top-5 $\downarrow$" denotes the Top-5 accuracy drop between pruned model and the baseline model. Other labels have the same meaning with Table \ref{table-res56-cifar10}.} 
    \label{table-res50-imagenet}
    \vspace{-0.6cm}
	    \begin{center}
	    \begin{threeparttable}
		    \begin{small}
			
			\begin{tabular}{l|ccccccc}
				\toprule[1.5pt]
				 Method 							&\makecell[c]{Base Top-1 \\  (\%)}	&\makecell[c]{Base Top-5 \\ (\%)}	&\makecell[c]{Pruned Top-1 \\ (\%)}	&\makecell[c]{Pruned Top-5 \\ (\%)}	& \makecell[c]{Top-1 $\downarrow$ \\ (\%)}	&	\makecell[c]{Top-5 $\downarrow$ \\ (\%)}	& 	\makecell[c]{FLOPs $\downarrow$ \\ (\%)}	
				
				\\  \midrule[1pt] 
				SFP \cite{he2018soft}	    &	76.15 	&	92.87	&	74.61	&	92.06	&	1.54	&	0.81 	&	41.8	\\
				GAL-0.5 \cite{lin2019towards}				&	76.15	&	92.87	&	71.95	&	90.94	&	4.20	&	1.93	&	43.03	\\
				HRank \cite{lin2020hrank}				&	76.15	&	92.87	&	74.98	&	92.33	&	1.17	&	0.54	&	43.76	\\
				NISP \cite{yu2018nisp}			&	-		&	-		&	-		&	-		&	0.89	&	-   	&	44.01	\\	
				Hinge \cite{li2020group}             &     -     &    -      &   74.7    &    -     &     -     &       &  46.55           \\
				Channel Pr \cite{he2017channel}&	- 		&	92.2	&	-		&	90.8	&	- 		&	1.4 	&	50		\\
				HP \cite{xu2018hybrid}			&	76.01 	&	92.93	&	74.87	&	92.43	&	1.14 	&	0.50 	&	50		\\
				NS \cite{liu2017learning}          &  75.04    &    -      &   69.60   &    -      &    5.44   &  -        &   50.51   \\
				MetaPruning \cite{liu2019metapruning}	&	76.6	&	-		&	75.4	&	-		&	1.2		&	-		&	51.10	\\
				Autopr \cite{luo2020autopruner}&	76.15 	&	92.87	&	74.76	&	92.15	&	1.39	&	0.72 	&	51.21	\\
				GDP \cite{lin2018accelerating}	&	75.13 	&	92.30	&	71.89	&	90.71	&	3.24 	&	1.59 	&	51.30	\\
				\textbf{SNF (ours)}	&	\textbf{76.75}	&	\textbf{93.15}	&	\textbf{76.01}	&	\textbf{92.83}	&	\textbf{0.74}	&	\textbf{0.32}	&	52.10	\\
				FPGM \cite{he2019filter}				&	76.15	&	92.87	&	74.83	&	92.32	&	1.32	&	0.55	&	53.5	\\
				DCP \cite{zhuang2018discrimination}&76.01 	&	92.93	&	74.95	&	92.32	&	1.06 	&	0.61 	&	55.76	\\		
				C-SGD \cite{ding2019centripetal}				&	75.33	&	92.56	&	74.54	&	92.09	&	0.79 	&	0.47 	&	55.76	\\
				ThiNet \cite{luo2017thinet}&	75.30	&	92.20	&	72.03	&	90.99	&	3.27 	&	1.21 	&	55.83	\\
				SASL \cite{shi2020sasl}				&	76.15	&	92.87	&	75.15	&	92.47	&	1.00	&	0.40	&	56.10	\\
				TRP \cite{xu2020trp}					&	75.90	&	92.70	&	72.69	&	91.41	&	3.21	&	1.29	&	56.52	\\
				LFPC \cite{he2020learning}		&	76.15	&	92.87	&	74.46	&	92.32	&	1.69	&	0.55	&	60.8	\\
				

				HRank \cite{lin2020hrank}	   &	76.15	&	92.87	&	71.98	&	91.01	&	4.17	&	1.86	&	62.10	\\
				\textbf{SNF(ours)}	&	\textbf{76.75}	&	\textbf{93.15}	&	\textbf{75.14}	&	\textbf{92.49}	&	\textbf{1.61}	&	\textbf{0.66}	&	62.10	\\
				ResRep \cite{ding2021resrep}    &   76.15   &   92.87   &   75.30   &   92.47   &   0.85    &0.40    &   62.10   \\
				
				\bottomrule[1.5pt]
			\end{tabular}
		\end{small}
		\end{threeparttable} 
	\end{center}
	\label{exp-table-imgnet}
\end{table*}

\textbf{Pruning ResNet-50 on ImageNet.}
As shown in Table \ref{table-res50-imagenet}, we compare our approach with other previous methods which once achieved the state-of-the-art performance using ResNet-50 on ImageNet. We train ResNet-50 from scratch and achieve the Top-1 accuracy of 76.75\% and Top-5 accuracy of 93.15\%. To our best knowledge, the accuracy of the baseline model is higher than the initial training results reported by other literature. We should know that pruning a higher accuracy model with a lossless accuracy is more challenging. 
For a fair comparison, we evaluate our method at different pruning rates of 52.10\% and 62.10\% respectively.
In Table \ref{table-res50-imagenet}, SNF achieves 76.01\% Top-1 accuracy and 92.83\% Top-5 accuracy when reducing 52.10\% FLOPs. The pruned model with 62.10\% FLOPs reduction loses 1.61\% Top-1 accuracy and 0.66\% Top-5 accuracy. Compared with the \cite{he2018soft, lin2019towards, lin2020hrank, yu2018nisp, li2020group, he2017channel, xu2018hybrid, liu2017learning, luo2020autopruner, lin2018accelerating}, we achieve the best accuracy (0.74\% and 0.32\% drop for Top-1 and Top-5 accuracy) with higher FLOPs reduction (52.10\%). Under the same FLOPs reduction(62.10\%), our method outperforms HRank \cite{lin2020hrank} by 2.56\%, but is worse than ResRep \cite{ding2021resrep} (SNF outperforms ResRep when using ResNet-56/110 on CIFAR-10).  Table \ref{table-res50-imagenet} shows that the model performance after pruning with our method defeats other methods when the pruning ratio is smaller than 52.10\%, and it is quite competitive at the higher pruning rates.

\begin{figure*}
  \centering
  \begin{subfigure}{0.25\linewidth}
    \centering
    \includegraphics[height=1in, width=1in]{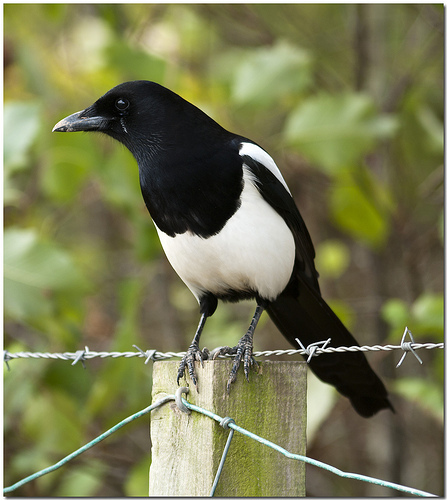}
    \caption{}
    \label{fig:vis-a}
  \end{subfigure}
  \hspace{-0.2in}
  \begin{subfigure}{0.25\linewidth}
    \centering
    \includegraphics[height=1in]{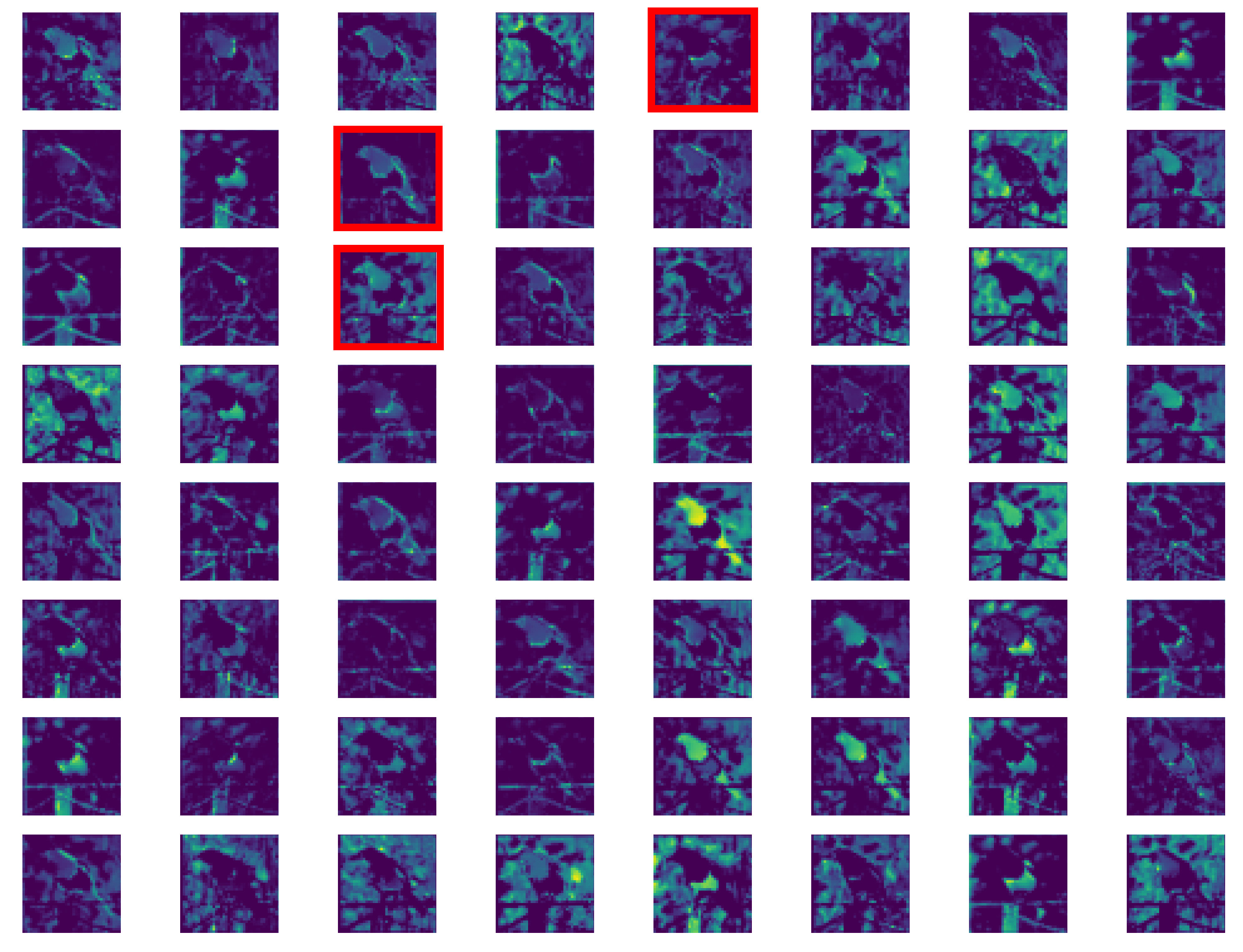}
    \caption{}
    \label{fig:vis-b}
  \end{subfigure}
  \hspace{-0.2in}
  \begin{subfigure}{0.25\linewidth}
    \centering
    \includegraphics[height=1in]{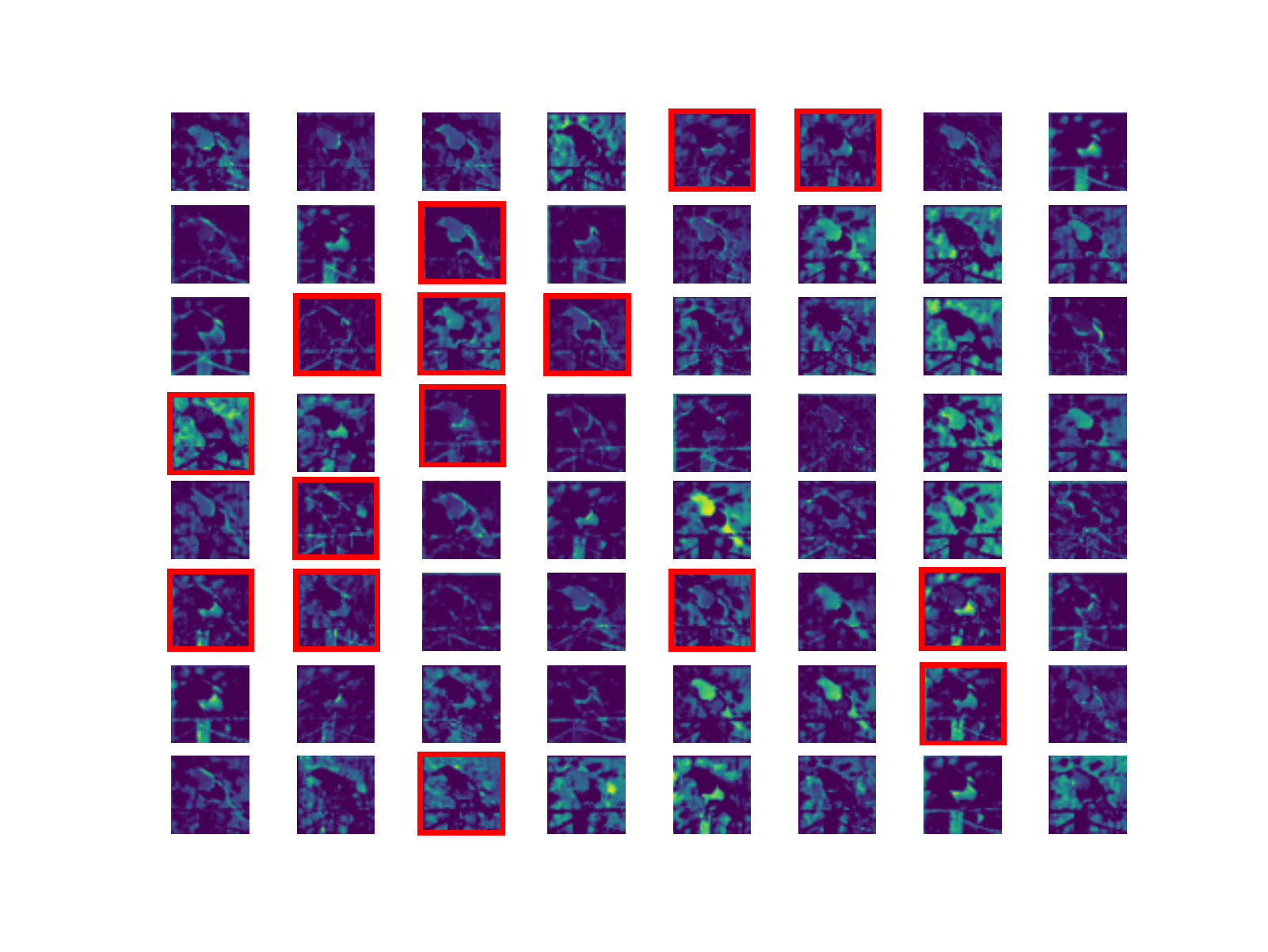}
    \caption{}
    \label{fig:vis-c}
  \end{subfigure}
  \hspace{-0.2in}
  \begin{subfigure}{0.25\linewidth}
    \centering
    \includegraphics[height=1in]{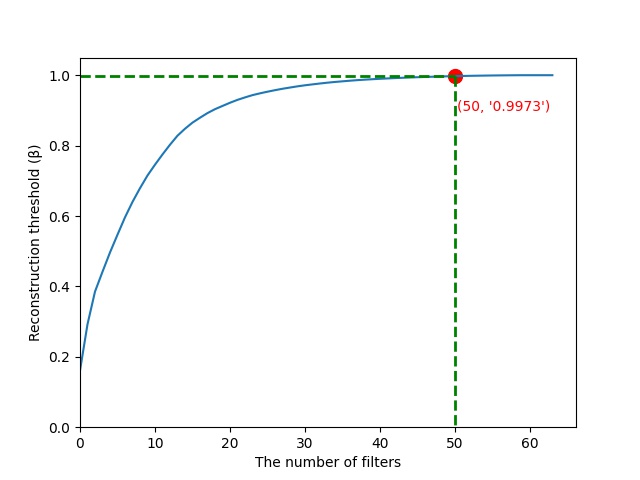}
    \caption{}
    \label{fig:vis-d}
  \end{subfigure}
  \caption{Visualizing the $\mathbf{3 \times 3}$ convolutional layer of the first block in ResNet-50. (a) Original picture. It is the input of ResNet-50. (b) Uniform Pruning. At the 5\% pruning ratio, there are 3 filters corresponding to the feature maps marked in red that will be removed. (c) SNF Pruning. The reconstruction threshold is 99.73\% at 5\% FLOPs reduction. There are 14 filters corresponding to the feature maps marked in red that will be removed. (d) Reconstruction threshold - The number of filters. The curve describes the relationship between the reconstruction threshold and the number of reserved filters. }
  \label{fig:feature-vis}
\end{figure*}
\begin{table}[t]
    \caption{The results of selecting the number of reserved filters with different methods on ResNet-56. All the labels have the same meaning with Table \ref{table-res56-cifar10}.}
    \label{table-res56-select-number}
 	\vspace{-0.5cm}
	\setlength{\tabcolsep}{0.2em}
	\begin{center}
		\begin{tabular}{l|cccc}	
            \toprule[1.5pt]
			\makecell[l]{Method}  &\makecell[c]{Base Top-1 \\ (\%)}   &\makecell[c]{Top-1 $\downarrow$ \\(\%)}  &\makecell[c]{FLOPs $\downarrow$ \\ (\%)}\\
			\midrule[1pt]
			
			Baseline Model  &93.61   &0   &0 \\
			Uniform Pruning  &93.37    &0.24  &52.42 \\
			Random Pruning  &93.41  &0.20   &52.44 \\
			\textbf{SNF Pruning}  &\textbf{93.75}   &\textbf{-0.14}   &\textbf{52.94} \\
			\midrule[1pt]
			ResRep \cite{ding2021resrep}  &-   &1.29  &77.83 \\
			ResRep* Pruning  &92.32   &1.29  &77.85 \\
			\textbf{SNF Pruning}  &\textbf{92.79}   &\textbf{0.82}  &\textbf{77.93} \\
			\bottomrule[1.5pt]
		\end{tabular}
		\begin{tablenotes}
			 \item * We only use the structure pruned by ResRep \cite{ding2021resrep}.
			\end{tablenotes}
		\vspace{-0.5cm}	
	\end{center}
\end{table}

\begin{figure}[t]
  \centering
  \vspace{-0.3cm}
   \includegraphics[trim=0 0 0 0,scale=0.5]{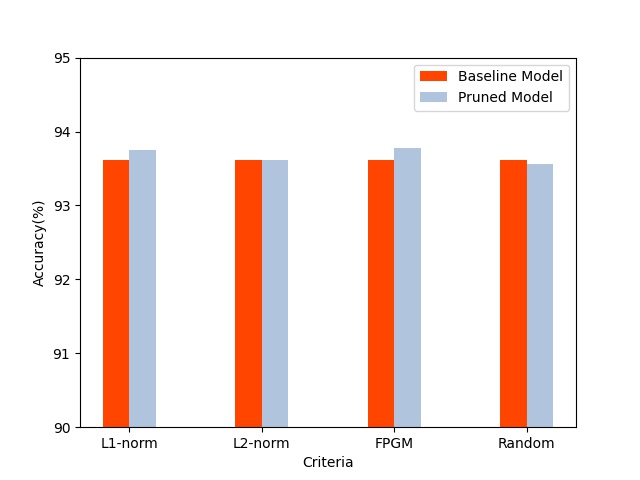}
   \caption{Accuracy - Criteria. The red columns are represented as the accuracy of the baseline model. The light steel blue columns are the accuracy of the pruned model which prunes with different criteria. 
}
   \label{fig:dif-criteria}
\end{figure}

\subsection{Ablation Study}
\textbf{Different Criteria for Filter Selection.}
The criterion is one of the necessary components for our proposed method. The criterion based on the magnitude of channel weights consists of $L_1$-norm and $L_2$-norm \cite{he2017channel}. These approaches are early used to sort filter importance. FPGM \cite{he2019filter} compresses models by pruning redundant filters with geometric median criteria. For analyzing the criteria of these methods fairly, we compare them on the ResNet-56 network at the same FLOPs. 

In Figure \ref{fig:dif-criteria}, it shows that the model performance of selecting filters with different criteria. For ResNet-56 on CIFAR-10 at 52.94\% pruning ratio, we can find that using the different criteria of filter selection has a close model performance. 
When pruning model with the criteria of $L1$-norm, $L2$-norm and FPGM, the accuracy of the pruned model is 93.75\%, 93.61\%, 93.78\% respectively. Compared with the performance of the model trained from the scratch, the accuracy of the pruned model has no obvious changes and they all have no accuracy drop. However, pruning filters with $L_p$-norm or FPGM has a significant improvement compared to random pruning. As shown in Figure \ref{fig:dif-criteria}, the accuracy drop obviously with random pruning. We believe a reasonable filter selection criteria can reduce the accuracy drop of pruned model effectively. 

\textbf{Different Approaches for Selecting the Number of Filters.}
We introduce that the number of filters reserved in each layer is important for the accuracy of the pruned model. For evaluating our point, we compare the performance with different methods which select the number of filters on ResNet-56. The results are shown in Table \ref{table-res56-select-number}. These methods are used to select the number of filters. For a fair comparison, all experiments adopt the $L_1$-norm criteria for filter importance ranking. The approaches include (1) Uniform Pruning. All the layers use the same pruning ratio. (2) Random Pruning. All the layers select the number of filters randomly. (3) ResRep* Pruning. After pruning the model in terms of ResRep \cite{ding2021resrep}, we record the number of filters in each layer. In other words, we finally get a structure ResRep* pruned by ResRep. And then we prune the initial training model with $L_1$-norm criteria according to the ResRep*. (4) SNF Pruning (Our method). Searching for the proper number of filters automatically.

As shown in Table \ref{table-res56-select-number}, it shows the results of pruned ResNet-56 model using different approaches for selecting the number of filters in each layer. The baseline model means training ResNet-56 from scratch.
We firstly compare ResRep \cite{ding2021resrep} with ResRep* Pruning, the Top-1 accuracy drop of ResRep* pruning is the same as ResRep when the FLOPs reduction is similar (77.83\% $vs$ 77.85\%). This result gives evidence that the number of reserved filters is critical for filter pruning. Compared with Uniform pruning and Random pruning, the Top-1 accuracy of our method is better than them at the slightly higher FLOPs reduction (93.75\% $vs$ 93.37\% and 93.41\%). Compared to ResRep* pruning, the number of filters selected by our method outperforms by 0.47\% at the similar FLOPs reduction (77.93\% $vs$ 77.85\%). Furthermore, we can believe that our method can reserve the number of filters in each layer properly, and reduce redundancy effectively.

\begin{figure}[t]
  \centering
  \vspace{-0.6cm}
   \includegraphics[trim=0 0 0 0,scale=0.5]{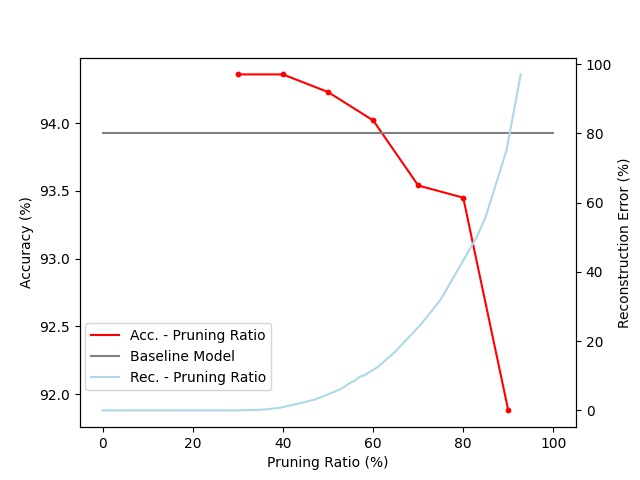}
   \caption{Acc. - Pruning Ratio (Red) and Rec. - Pruning Ratio (Light blue). "Acc." indicates the accuracy of the model. "Rec." denotes the reconstruction error. The accuracy and reconstruction error change with the pruning ratio for ResNet-110 on CIFAR-10. The gray line is the accuracy of ResNet-110 trained from scratch.
}
   \label{fig:dif-number}
\end{figure}

\textbf{Accuracy and Reconstruction Error Change with Pruning Ratio.}
As shown in Figure \ref{fig:dif-number}, to prove the stability of our method, we estimate the Top-1 accuracy at different pruning rates on ResNet-110. As the pruning ratio increases, the accuracy of the pruned model is higher than the baseline model before 68.68\% FLOPs reduction and then starts to drop quickly.
Few filters can not extract the feature sufficiently. It seems that all the related filters are removed at near 68.68\% FLOPs reduction, and the rest of the filters are irreplaceable. 
In Figure \ref{fig:dif-number}, we also report the reconstruction error threshold changes over the pruning rates. The reconstruction error (blue curve) starts to drop after the pruning ratio of 40\%, and the curve is getting steeper and steeper as the pruning ratio increases. More importantly, the accuracy decrease and reconstruction error threshold increase obviously are almost at the same pruning ratio. The result proves our method is effective and reasonable.

\textbf{Visualization.}
As shown in Figure \ref{fig:feature-vis}, we visualize the first $3 \times 3$ convolutional layer in the first block of ResNet-50. The pruned feature maps are marked with red boxes when reducing 5\% FLOPs. 
We label the sub-figure from left to right. 
(a) Original picture. It is the input of the ResNet-50. 
(b) Uniform Pruning. All the layers prune filters with the same pruning ratio. With the uniform pruning method, this layer will remove 3 filters. 
(c) SNF Pruning. Our method prunes 14 filters in this layer. 
(d) The reconstruction threshold - the number of filters reserved curve.
We further analyze by comparing Uniform Pruning with SNF Pruning. This layer has 64 filters corresponding to 64 feature maps. As shown in sub-figure (b) and (c), many different filters extract similar feature maps for the same input picture. There are some weakly correlated filters in this layer. Moreover, the content richness of feature maps is almost the same, and the ability of filter representation is close. The number of reserved filters with Uniform Pruning and SNF Pruning varies greatly. We have to consider whether pruning 14 filters at 5\% FLOPs reduction is reasonable. 
From the sub-figure (d), we can find the curve gradually becomes smoother as the number of reserved filters increases. The curve shows that the number of reserved filters is not sensitive when the reconstruction threshold is small. On the other hand, a few filters also can ensure the reconstruction error is very small and we can prune the filters boldly.

\section{Conclusion}
In this paper, we theoretically analyzed that the ability of filter representation is close to each other in the same layer. We naturally considered that the performance of the pruned model is limited to the number of filters reserved in each layer. Based on this idea, we proposed a new method called SNF, which consists of searching for the number of filters reserved in each layer and sorting the importance of filters. 
To search for the proper number of reserved filters, we thought of it as an optimization problem. The optimization goal is to minimize the reconstruction error between the original filters and the reduced dimensionality filters.
For sorting the importance of filters, we chose existing criteria to judge the filter importance.
The results of the experiment have shown that our method achieved the state-of-the-art performance on CIFAR-10 and attained the competitive performance on ImageNet. We believe our approach can give new inspiration for the filter pruning tasks.


{\small
\bibliographystyle{ieee_fullname}
\bibliography{egbib}

\begin{thebibliography}{10}\itemsep=-1pt

\bibitem{ba2013deep}
Lei~Jimmy Ba and Rich Caruana.
\newblock Do deep nets really need to be deep?
\newblock {\em arXiv preprint arXiv:1312.6184}, 2013.

\bibitem{banner2018post}
Ron Banner, Yury Nahshan, Elad Hoffer, and Daniel Soudry.
\newblock Post-training 4-bit quantization of convolution networks for
  rapid-deployment.
\newblock {\em arXiv preprint arXiv:1810.05723}, 2018.

\bibitem{chollet2017xception}
Fran{\c{c}}ois Chollet.
\newblock Xception: Deep learning with depthwise separable convolutions.
\newblock In {\em Proceedings of the IEEE conference on computer vision and
  pattern recognition}, pages 1251--1258, 2017.

\bibitem{cordts2016cityscapes}
Marius Cordts, Mohamed Omran, Sebastian Ramos, Timo Rehfeld, Markus Enzweiler,
  Rodrigo Benenson, Uwe Franke, Stefan Roth, and Bernt Schiele.
\newblock The cityscapes dataset for semantic urban scene understanding.
\newblock In {\em Proceedings of the IEEE conference on computer vision and
  pattern recognition}, pages 3213--3223, 2016.

\bibitem{deng2009imagenet}
Jia Deng, Wei Dong, Richard Socher, Li-Jia Li, Kai Li, and Li Fei-Fei.
\newblock Imagenet: A large-scale hierarchical image database.
\newblock In {\em 2009 IEEE conference on computer vision and pattern
  recognition}, pages 248--255. Ieee, 2009.

\bibitem{denil2013predicting}
Misha Denil, Babak Shakibi, Laurent Dinh, Marc'Aurelio Ranzato, and Nando
  De~Freitas.
\newblock Predicting parameters in deep learning.
\newblock {\em arXiv preprint arXiv:1306.0543}, 2013.

\bibitem{denton2014exploiting}
Emily~L Denton, Wojciech Zaremba, Joan Bruna, Yann LeCun, and Rob Fergus.
\newblock Exploiting linear structure within convolutional networks for
  efficient evaluation.
\newblock In {\em Advances in neural information processing systems}, pages
  1269--1277, 2014.

\bibitem{ding2019centripetal}
Xiaohan Ding, Guiguang Ding, Yuchen Guo, and Jungong Han.
\newblock Centripetal sgd for pruning very deep convolutional networks with
  complicated structure.
\newblock In {\em Proceedings of the IEEE/CVF Conference on Computer Vision and
  Pattern Recognition}, pages 4943--4953, 2019.

\bibitem{ding2019global}
Xiaohan Ding, Guiguang Ding, Xiangxin Zhou, Yuchen Guo, Jungong Han, and Ji
  Liu.
\newblock Global sparse momentum sgd for pruning very deep neural networks.
\newblock {\em arXiv preprint arXiv:1909.12778}, 2019.

\bibitem{ding2021resrep}
Xiaohan Ding, Tianxiang Hao, Jianchao Tan, Ji Liu, Jungong Han, Yuchen Guo, and
  Guiguang Ding.
\newblock Resrep: Lossless cnn pruning via decoupling remembering and
  forgetting.
\newblock In {\em Proceedings of the IEEE/CVF International Conference on
  Computer Vision}, pages 4510--4520, 2021.

\bibitem{frankle2018lottery}
Jonathan Frankle and Michael Carbin.
\newblock The lottery ticket hypothesis: Finding sparse, trainable neural
  networks.
\newblock {\em arXiv preprint arXiv:1803.03635}, 2018.

\bibitem{gao2021network}
Shangqian Gao, Feihu Huang, Weidong Cai, and Heng Huang.
\newblock Network pruning via performance maximization.
\newblock In {\em Proceedings of the IEEE/CVF Conference on Computer Vision and
  Pattern Recognition}, pages 9270--9280, 2021.

\bibitem{guo2016dynamic}
Yiwen Guo, Anbang Yao, and Yurong Chen.
\newblock Dynamic network surgery for efficient dnns.
\newblock {\em arXiv preprint arXiv:1608.04493}, 2016.

\bibitem{han2015deep}
Song Han, Huizi Mao, and William~J Dally.
\newblock Deep compression: Compressing deep neural networks with pruning,
  trained quantization and huffman coding.
\newblock {\em arXiv preprint arXiv:1510.00149}, 2015.

\bibitem{han2015learning}
Song Han, Jeff Pool, John Tran, and William~J Dally.
\newblock Learning both weights and connections for efficient neural networks.
\newblock {\em arXiv preprint arXiv:1506.02626}, 2015.

\bibitem{he2016deep}
Kaiming He, Xiangyu Zhang, Shaoqing Ren, and Jian Sun.
\newblock Deep residual learning for image recognition.
\newblock In {\em Proceedings of the IEEE conference on computer vision and
  pattern recognition}, pages 770--778, 2016.

\bibitem{he2020learning}
Yang He, Yuhang Ding, Ping Liu, Linchao Zhu, Hanwang Zhang, and Yi Yang.
\newblock Learning filter pruning criteria for deep convolutional neural
  networks acceleration.
\newblock In {\em Proceedings of the IEEE/CVF conference on computer vision and
  pattern recognition}, pages 2009--2018, 2020.

\bibitem{he2018soft}
Yang He, Guoliang Kang, Xuanyi Dong, Yanwei Fu, and Yi Yang.
\newblock Soft filter pruning for accelerating deep convolutional neural
  networks.
\newblock {\em arXiv preprint arXiv:1808.06866}, 2018.

\bibitem{he2018amc}
Yihui He, Ji Lin, Zhijian Liu, Hanrui Wang, Li-Jia Li, and Song Han.
\newblock Amc: Automl for model compression and acceleration on mobile devices.
\newblock In {\em Proceedings of the European conference on computer vision
  (ECCV)}, pages 784--800, 2018.

\bibitem{he2019filter}
Yang He, Ping Liu, Ziwei Wang, Zhilan Hu, and Yi Yang.
\newblock Filter pruning via geometric median for deep convolutional neural
  networks acceleration.
\newblock In {\em Proceedings of the IEEE/CVF Conference on Computer Vision and
  Pattern Recognition}, pages 4340--4349, 2019.

\bibitem{he2017channel}
Yihui He, Xiangyu Zhang, and Jian Sun.
\newblock Channel pruning for accelerating very deep neural networks.
\newblock In {\em Proceedings of the IEEE international conference on computer
  vision}, pages 1389--1397, 2017.

\bibitem{hinton2015distilling}
Geoffrey Hinton, Oriol Vinyals, and Jeff Dean.
\newblock Distilling the knowledge in a neural network.
\newblock {\em arXiv preprint arXiv:1503.02531}, 2015.

\bibitem{howard2017mobilenets}
Andrew~G Howard, Menglong Zhu, Bo Chen, Dmitry Kalenichenko, Weijun Wang,
  Tobias Weyand, Marco Andreetto, and Hartwig Adam.
\newblock Mobilenets: Efficient convolutional neural networks for mobile vision
  applications.
\newblock {\em arXiv preprint arXiv:1704.04861}, 2017.

\bibitem{hu2016network}
Hengyuan Hu, Rui Peng, Yu-Wing Tai, and Chi-Keung Tang.
\newblock Network trimming: A data-driven neuron pruning approach towards
  efficient deep architectures.
\newblock {\em arXiv preprint arXiv:1607.03250}, 2016.

\bibitem{jaderberg2014speeding}
Max Jaderberg, Andrea Vedaldi, and Andrew Zisserman.
\newblock Speeding up convolutional neural networks with low rank expansions.
\newblock {\em arXiv preprint arXiv:1405.3866}, 2014.

\bibitem{krizhevsky2009learning}
Alex Krizhevsky, Geoffrey Hinton, et~al.
\newblock Learning multiple layers of features from tiny images.
\newblock 2009.

\bibitem{lee2018snip}
Namhoon Lee, Thalaiyasingam Ajanthan, and Philip~HS Torr.
\newblock Snip: Single-shot network pruning based on connection sensitivity.
\newblock {\em arXiv preprint arXiv:1810.02340}, 2018.

\bibitem{li2016pruning}
Hao Li, Asim Kadav, Igor Durdanovic, Hanan Samet, and Hans~Peter Graf.
\newblock Pruning filters for efficient convnets.
\newblock {\em arXiv preprint arXiv:1608.08710}, 2016.

\bibitem{li2020group}
Yawei Li, Shuhang Gu, Christoph Mayer, Luc~Van Gool, and Radu Timofte.
\newblock Group sparsity: The hinge between filter pruning and decomposition
  for network compression.
\newblock In {\em Proceedings of the IEEE/CVF conference on computer vision and
  pattern recognition}, pages 8018--8027, 2020.

\bibitem{lin2020hrank}
Mingbao Lin, Rongrong Ji, Yan Wang, Yichen Zhang, Baochang Zhang, Yonghong
  Tian, and Ling Shao.
\newblock Hrank: Filter pruning using high-rank feature map.
\newblock In {\em Proceedings of the IEEE/CVF Conference on Computer Vision and
  Pattern Recognition}, pages 1529--1538, 2020.

\bibitem{lin2018accelerating}
Shaohui Lin, Rongrong Ji, Yuchao Li, Yongjian Wu, Feiyue Huang, and Baochang
  Zhang.
\newblock Accelerating convolutional networks via global \& dynamic filter
  pruning.
\newblock In {\em IJCAI}, volume~2, page~8, 2018.

\bibitem{lin2019towards}
Shaohui Lin, Rongrong Ji, Chenqian Yan, Baochang Zhang, Liujuan Cao, Qixiang
  Ye, Feiyue Huang, and David Doermann.
\newblock Towards optimal structured cnn pruning via generative adversarial
  learning.
\newblock In {\em Proceedings of the IEEE/CVF Conference on Computer Vision and
  Pattern Recognition}, pages 2790--2799, 2019.

\bibitem{lin2014microsoft}
Tsung-Yi Lin, Michael Maire, Serge Belongie, James Hays, Pietro Perona, Deva
  Ramanan, Piotr Doll{\'a}r, and C~Lawrence Zitnick.
\newblock Microsoft coco: Common objects in context.
\newblock In {\em European conference on computer vision}, pages 740--755.
  Springer, 2014.

\bibitem{liu2019structured}
Yifan Liu, Ke Chen, Chris Liu, Zengchang Qin, Zhenbo Luo, and Jingdong Wang.
\newblock Structured knowledge distillation for semantic segmentation.
\newblock In {\em Proceedings of the IEEE/CVF Conference on Computer Vision and
  Pattern Recognition}, pages 2604--2613, 2019.

\bibitem{liu2017learning}
Zhuang Liu, Jianguo Li, Zhiqiang Shen, Gao Huang, Shoumeng Yan, and Changshui
  Zhang.
\newblock Learning efficient convolutional networks through network slimming.
\newblock In {\em Proceedings of the IEEE international conference on computer
  vision}, pages 2736--2744, 2017.

\bibitem{liu2019metapruning}
Zechun Liu, Haoyuan Mu, Xiangyu Zhang, Zichao Guo, Xin Yang, Kwang-Ting Cheng,
  and Jian Sun.
\newblock Metapruning: Meta learning for automatic neural network channel
  pruning.
\newblock In {\em Proceedings of the IEEE/CVF International Conference on
  Computer Vision}, pages 3296--3305, 2019.

\bibitem{liu2018bi}
Zechun Liu, Baoyuan Wu, Wenhan Luo, Xin Yang, Wei Liu, and Kwang-Ting Cheng.
\newblock Bi-real net: Enhancing the performance of 1-bit cnns with improved
  representational capability and advanced training algorithm.
\newblock In {\em Proceedings of the European conference on computer vision
  (ECCV)}, pages 722--737, 2018.

\bibitem{long2015fully}
Jonathan Long, Evan Shelhamer, and Trevor Darrell.
\newblock Fully convolutional networks for semantic segmentation.
\newblock In {\em Proceedings of the IEEE conference on computer vision and
  pattern recognition}, pages 3431--3440, 2015.

\bibitem{luo2020autopruner}
Jian-Hao Luo and Jianxin Wu.
\newblock Autopruner: An end-to-end trainable filter pruning method for
  efficient deep model inference.
\newblock {\em Pattern Recognition}, 107:107461, 2020.

\bibitem{luo2017thinet}
Jian-Hao Luo, Jianxin Wu, and Weiyao Lin.
\newblock Thinet: A filter level pruning method for deep neural network
  compression.
\newblock In {\em Proceedings of the IEEE international conference on computer
  vision}, pages 5058--5066, 2017.

\bibitem{molchanov2019importance}
Pavlo Molchanov, Arun Mallya, Stephen Tyree, Iuri Frosio, and Jan Kautz.
\newblock Importance estimation for neural network pruning.
\newblock In {\em Proceedings of the IEEE/CVF Conference on Computer Vision and
  Pattern Recognition}, pages 11264--11272, 2019.

\bibitem{molchanov2016pruning}
Pavlo Molchanov, Stephen Tyree, Tero Karras, Timo Aila, and Jan Kautz.
\newblock Pruning convolutional neural networks for resource efficient
  inference.
\newblock {\em arXiv preprint arXiv:1611.06440}, 2016.

\bibitem{radosavovic2019network}
Ilija Radosavovic, Justin Johnson, Saining Xie, Wan-Yen Lo, and Piotr
  Doll{\'a}r.
\newblock On network design spaces for visual recognition.
\newblock In {\em Proceedings of the IEEE/CVF International Conference on
  Computer Vision}, pages 1882--1890, 2019.

\bibitem{ren2015faster}
Shaoqing Ren, Kaiming He, Ross Girshick, and Jian Sun.
\newblock Faster r-cnn: Towards real-time object detection with region proposal
  networks.
\newblock {\em Advances in neural information processing systems}, 28:91--99,
  2015.

\bibitem{russakovsky2015imagenet}
Olga Russakovsky, Jia Deng, Hao Su, Jonathan Krause, Sanjeev Satheesh, Sean Ma,
  Zhiheng Huang, Andrej Karpathy, Aditya Khosla, Michael Bernstein, et~al.
\newblock Imagenet large scale visual recognition challenge.
\newblock {\em International journal of computer vision}, 115(3):211--252,
  2015.

\bibitem{shi2020sasl}
Jun Shi, Jianfeng Xu, Kazuyuki Tasaka, and Zhibo Chen.
\newblock Sasl: saliency-adaptive sparsity learning for neural network
  acceleration.
\newblock {\em IEEE Transactions on Circuits and Systems for Video Technology},
  31(5):2008--2019, 2020.

\bibitem{wang2021convolutional}
Zi Wang, Chengcheng Li, and Xiangyang Wang.
\newblock Convolutional neural network pruning with structural redundancy
  reduction.
\newblock In {\em Proceedings of the IEEE/CVF Conference on Computer Vision and
  Pattern Recognition}, pages 14913--14922, 2021.

\bibitem{xu2018hybrid}
Xiaofan Xu, Mi~Sun Park, and Cormac Brick.
\newblock Hybrid pruning: Thinner sparse networks for fast inference on edge
  devices.
\newblock {\em arXiv preprint arXiv:1811.00482}, 2018.

\bibitem{xu2020trp}
Yuhui Xu, Yuxi Li, Shuai Zhang, Wei Wen, Botao Wang, Yingyong Qi, Yiran Chen,
  Weiyao Lin, and Hongkai Xiong.
\newblock Trp: Trained rank pruning for efficient deep neural networks.
\newblock {\em arXiv preprint arXiv:2004.14566}, 2020.

\bibitem{ye2018rethinking}
Jianbo Ye, Xin Lu, Zhe Lin, and James~Z Wang.
\newblock Rethinking the smaller-norm-less-informative assumption in channel
  pruning of convolution layers.
\newblock {\em arXiv preprint arXiv:1802.00124}, 2018.

\bibitem{yim2017gift}
Junho Yim, Donggyu Joo, Jihoon Bae, and Junmo Kim.
\newblock A gift from knowledge distillation: Fast optimization, network
  minimization and transfer learning.
\newblock In {\em Proceedings of the IEEE Conference on Computer Vision and
  Pattern Recognition}, pages 4133--4141, 2017.

\bibitem{yu2018nisp}
Ruichi Yu, Ang Li, Chun-Fu Chen, Jui-Hsin Lai, Vlad~I Morariu, Xintong Han,
  Mingfei Gao, Ching-Yung Lin, and Larry~S Davis.
\newblock Nisp: Pruning networks using neuron importance score propagation.
\newblock In {\em Proceedings of the IEEE Conference on Computer Vision and
  Pattern Recognition}, pages 9194--9203, 2018.

\bibitem{zhuang2018discrimination}
Zhuangwei Zhuang, Mingkui Tan, Bohan Zhuang, Jing Liu, Yong Guo, Qingyao Wu,
  Junzhou Huang, and Jinhui Zhu.
\newblock Discrimination-aware channel pruning for deep neural networks.
\newblock {\em arXiv preprint arXiv:1810.11809}, 2018.

\end{thebibliography}
}

\end{document}